\documentclass{article}

    \usepackage[preprint]{neurips_2025}

\usepackage[utf8]{inputenc} 
\usepackage[T1]{fontenc}    
\usepackage{hyperref}       
\usepackage{url}            
\usepackage{booktabs}       
\usepackage{amsfonts}       
\usepackage{nicefrac}       
\usepackage{microtype}      
\usepackage{xcolor}         

\usepackage{mathtools}
\usepackage{amssymb}
\usepackage{subcaption}
\usepackage{algorithm2e}
\usepackage{placeins}
\title{Linearity-based neural network compression}

\author{
Silas Dobler\\
University of Passau, Germany\\
Silas.dobler@gmx.de\\
\And
Florian Lemmerich\\
University of Passau, Germany\\
Florian.lemmerich@uni-passau.de\\
}

\begin{document}

\maketitle

\begin{abstract}
In neural network compression, most current methods reduce unnecessary parameters by measuring importance and redundancy.
To augment already highly optimized existing solutions, we propose \emph{linearity-based compression} as a novel way to reduce weights in a neural network.
It is based on the intuition that with ReLU-like activation functions, neurons that are almost always activated behave linearly, allowing for merging of subsequent layers.
We introduce the theory underlying this compression and evaluate our approach experimentally.
Our novel method achieves a lossless compression down to 1/4 of the original model size in over the majority of tested models. 
Applying our method on already importance-based pruned models shows very little interference between different types of compression, demonstrating the option of successful combination of techniques.  
Overall, our work lays the foundation for 
a new type of compression method that enables smaller and ultimately more efficient neural network models.
\end{abstract}

\section{Introduction}

%
The number of parameters in deep neural networks has dramatically grown in the last decade, raising fundamental concerns about the computational and thus indirectly environmental costs.
Concurrently, use cases of the deep learning models also expanded to systems with restricted computational power or memory, such as mobile phones or embedded systems.\citep{MobileNets}
To reduce costs and enable distributed usage of deep learning models, network compression attempts to reduce the size of existing models to a manageable size with minimal functional impact.\citep{Overview_network_compression}
In this direction, several highly optimized ways to reduce model size have been proposed, including quantization, removing unimportant weights or combining redundant weights. State-of-the-art methods combine best efforts of all approaches to achieve the highest compression while maintaining as much performance as possible.\citep{deep_compression}
The main objective of this work is to introduce a fundamentally new approach for compressing neural networks based on \emph{linearity}. Linearity of an activation function allows redistributing weights and removing the neuron itself. Linear neurons are the most active neurons and therefore not widely considered for compression in other approaches. An up-following research question addresses the compression performance of this approach in comparison with existing methods.

\begin{figure}[t]
    \centering
       \includegraphics[width=0.75\linewidth]{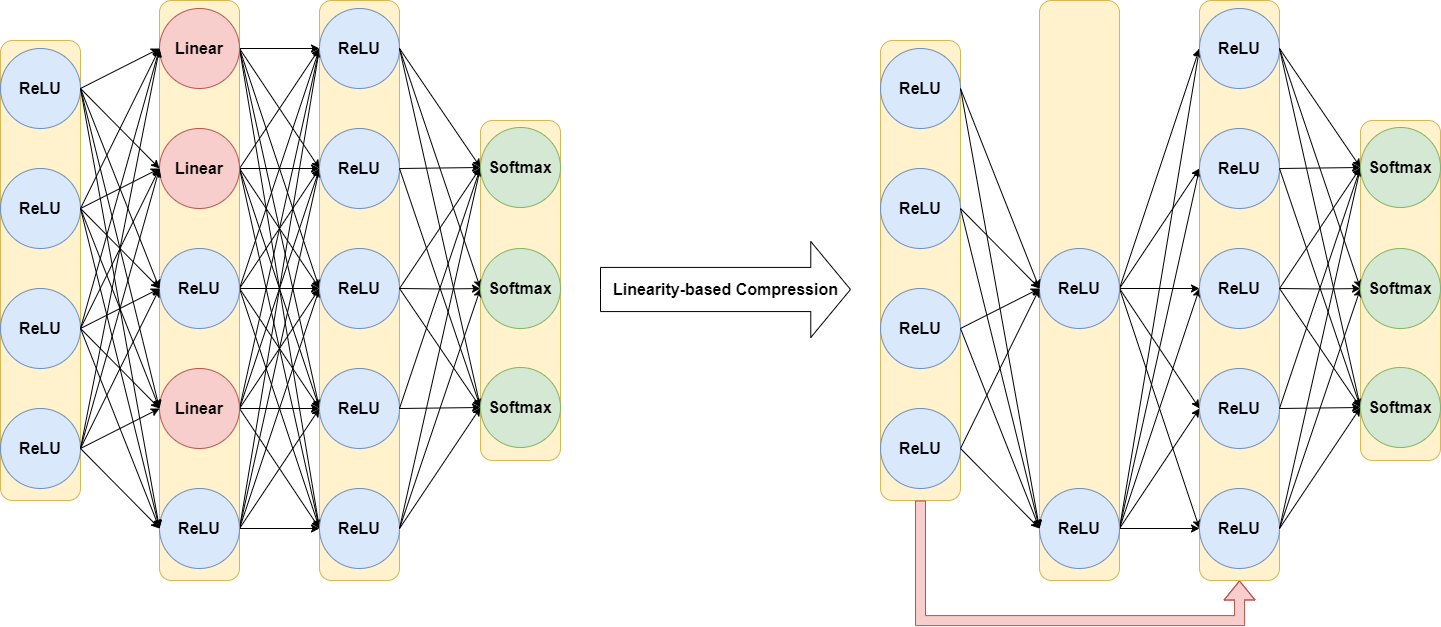}
    \caption{Example of linearity-based network compression. Replacing all linear neurons in the second layer of the original model with a shortcut connection between the neurons input and output layer.}
    \label{fig:paperConcept}
\end{figure}

The main idea of this work is to exploit the partial linearity of neurons with ReLU-like activations to multiply out their weights into the next layer.
We start with theoretical considerations regarding this idea, which allows reshaping the model to use fewer parameters with guaranteed identical outputs under certain conditions. Next, we demonstrate how these conditions can be relaxed to develop a practical approach that enables to prune more neurons but only approximates the original output. Figure \ref{fig:paperConcept} summarizes our approach: Neurons with (theoretically or empirically) linear activations only are replaced with direct connections between their input and output layer.
Finally, the compression performance of this prototype is experimentally evaluated on both a standard fully connected neural network and a network that has already been pruned using an alternative method.

The main contribution of this work is the introduction of a novel way of pruning neural networks. For this approach, we present a theoretical foundation that provides a justification for further compression methods based on linearity.
Furthermore, we demonstrate that our prototype algorithm exploiting this idea leads for many datasets to substantial compression in network size while maintaining prediction performance. The code of our implementation is available for others to adapt or build upon.\footnote{https://github.com/sdobler19/linearity-based-neural-network-compression}
%

\section{Mathematical Justification}
\label{section:basicIdea}
The idea underlying our proposed approach is to exploit the ineffectiveness of neurons with a linear activation function in the network.
We begin by demonstrating how a single linear neuron can be eliminated by appropriately redistributing its weights. Next, we explore how even with non-linear ReLU activation, certain neurons may behave linearly under specific conditions. Finally, we examine the impact of this approach on the overall size of the model.

\subsection{Removal of Linear Neurons}
As it is well known, in a standard feedforward neural network with exclusively linear activation functions, linear neurons can be collapsed into the next layer of the network by multiplying out the parameters.
As a result, the adjusted weights and biases in the following layer account for the inputs originally processed by the removed linear neurons.
The linear neuron is dissolved in the process.

We consider a fully connected neural network with k hidden layers, denoted $h_i$ for $i = 1, \ldots, k-1$, where each layer $h_i$ contains $m_i$ neurons. Assume that all but one neuron use a ReLU activation, and we denote their outputs $h_{ij}$, with associated weights $w^{(ij)}_{h_{lm}}$ and biases $b_{ij}$. 
As the exception, we let (without loss of generality) the first neuron of layer $i$ have a linear activation function $f(x)$, i.e., $f(x) = x$.
Then, the next layer's neurons then can be represented by
\begin{align}
    h_{i+1p} = ReLU((\sum_{j = 1}^{m_i} w^{(i+1p)}_{h_{ij}} h_{ij}) + b_{i+1p}) \quad \text{for } p = 1, ..., m_{i+1} \label{layer3}
\end{align}
Layer $h_i$ can be represented in the same way, with adaption to the linear first neuron.
\begin{align}
    &h_i = ((\sum_{j = 1}^{m_{i-1}} w^{(i-1, 1)}_{h_{i-1j}} h_{i-1j}) + b_{i-1,1}, 
    ReLU(...)
    , ..., ReLU((\sum_{j = 1}^{m_{i-1}} w^{(i-1, m_1)}_{h_{i-1j}} h_{i-1j}) + b_{i-1,m_i})) \label{layer2}
\end{align}
Inserting \ref{layer2} into \ref{layer3} and exploiting the linearity gives the following representation for the neurons in the next layer:
\begin{align}
    h_{i+1p} &= ReLU((w^{(i+1p)}_{h_{i1}} ((\sum_{l = 1}^{m_{i-1}} w^{(i1)}_{h_{i-1l}} h_{i-1l}) + b_{i1}) + \sum_{j = 2}^{m_i} w^{(ip)}_{h_{ij}} h_{ij}) + b_{i+1p})  \notag\\
    &= ReLU((\sum_{j = 2}^{m_i} w^{(ip)}_{h_{ij}} h_{ij}) + (\sum_{l = 1}^{m_{i-1}} w^{(i+1p)}_{h_{i1}} w^{(i1)}_{h_{i-1l}} h_{i-1l} ) + w^{(i+1p)}_{h_{i1}}b_{i1} +  b_{i+1p})  \notag\\
    &= ReLU((\sum_{j = 2}^{m_i} w^{(ip)}_{h_{ij}} h_{ij}) + (\sum_{l = 1}^{m_{i-1}} w^{(i1)}_{h_{i-1l}--updated} h_{i-1l} ) + b_{i+1p--updated})
    \label{composition_exploit}
\end{align}
for $p= 1, ..., m_{i+1}$. The weights $w^{(ij)}$ of a linear neuron $h_{ij}$ are updated by multiplying them by the weight corresponding to the output of the linear neuron in each neuron of the next layer $w^{(i+1p)}_{h_{i1}}$ before adding the resulting weights to those neurons. 
The same updating rule applies to the bias.
To ensure the network produces identical predictions, the inputs originally received by the linear neuron must also be provided directly to the neurons in the next layer. The linear neuron and all its corresponding weights can then be removed.
Now that we know that a linear neuron can be removed, we will see how it can happen that neurons with the non-linear activation of ReLUs still end up being linear.
\subsection{Linearity of ReLUs}
ReLU activation functions operate by outputting zero for negative inputs and applying the identity function $f(x) = x$ for non-negative inputs. The special setting in the neural network can lead to conditions where all possible inputs to a ReLU are non-negative.
Such a ReLU with restricted domain $ReLU:\mathbb{R}^+_0 \rightarrow \mathbb{R}^+_0$ allows the conclusion,
\begin{equation}
    ReLU(x) = max(0, x) = x 
\end{equation}
resulting in linear ReLUs, which could be removed as previously seen.

For a ReLU input in the neural network (f(x) = $\sum_{i = 1}^{n} x_i w_i + b$) to always be non-negative, two conditions must hold: all neuron inputs $x_i$ must be non-negative, and all weights $w_i$ and the bias must be non-negative. The first is guaranteed in deeper layers if all hidden layers use non-negative activation functions like ReLU. The second can be verified directly in a trained or pre-trained model. Thus, in a feedforward network with ReLU activations, any neuron from the second layer onward with non-negative weights and bias behaves linearly and can be removed.
\subsection{Combination of Multiple Linear Neurons}
The number of parameters associated with a single neuron in a fully connected network depends on the number of neurons in the previous layer $width_{i-1}$ and the number of neurons in the next layer $width_{i+1}$.
Without considering any unrelated parameters, the neuron initially carries one weight for each neuron in the previous layer and one bias. In addition, each neuron in the next layer carries one weight for the neuron we are examining. The total number of weights then adds up to $param_{original} = width_{i-1} + 1 + width_{i+1}$.
During compression, all of these parameters are removed. However, the distribution of weights into the next layer introduces some more parameters. Specifically, this redistribution fully connects the previous layer directly to the next layer, resulting in an additional $ param_{compr} = width_{i-1} width_{i + 1}$
weights.
If both the previous and next layers contain more than 3 neurons, the number of parameters added through this distribution exceeds the number originally removed. In modern models with thousands of neurons, such compression actually increases the total number of parameters.\citep{GPT3}

We now extend the analysis to a setting where multiple linear neurons are present in the current layer. Suppose there are $n_{linear}$ linear neurons. Each of these is originally associated with $width_{i-1} + 1 + width_{i+1}$ weights, resulting in 
\begin{equation}
    param_{original} = n_{linear} (width_{i-1} + 1 + width_{i+1})
\end{equation}
During compression, each of the $n_{linear}$ linear neurons redistributes its input weights into the next layer, creating connections from every neuron in the previous layer to every neuron in the next layer. Although multiple linear neurons contribute to these connections, the resulting weights for the same input-output pairs can be summed, effectively collapsing them into a single weight per connection. As a result, the total number of new parameters remains at $width_{i-1} width_{i + 1}$ regardless of how many linear neurons are compressed.
This leads to the general inequation 
\begin{align}
    & param_{compr} \leq param_{original}  \notag\\
    & width_{i-1} width_{i + 1} \leq n_{linear} (width_{i-1} + 1 + width_{i+1}) \label{equ:compressionImprov}
\end{align}
A sufficient number of linear neurons in a layer is needed to exploit the compression effectively.
However, in practice, it is unlikely that a large enough set of provably linear neurons will always be present. The next section therefore introduces a more practical method based on empirical linearity before testing the compression in experiments.
\section{Practical Prototype}
Next, we want to develop a practical compression approach based on this theory. It is based on approximating the linearity as well as the contours of the practical implementation. 
\subsection{Empirical Linearity and Activation Rates}
Previously, the work considered provably linear neurons that are always activated based on their input and parameters. 
However, there also exist neurons that are not provably linear but are nevertheless activated either consistently or at least very frequently across realistic data samples.
These neurons effectively behave linearly within the network, even if not guaranteed by the weight structure.
Our practical approach exploits the removal of such neurons.
The choice of how to measure linearity is, like identifying unimportance, a key degree of freedom.
We propose to measure linearity by the ratio of neuron activations during dataset inference.
To determine linearity, we query the model and analyze the behaviour of the neurons in an additional dataset we call the pruning set.
For this purpose, we observe for each input instance, which neurons are activated, i.e., if its output is greater than $0$. Averaging over the number (not the size) of individual activations for each neuron yields the activation rates. Those reflect the proportion of examples in which the neuron was active and serves as an empirical measure of its linearity.
By lowering the activation rate threshold used to determine which neurons are considered sufficiently linear for compression, one can pick in the trade-off between size and performance.
\subsection{Shortcut Connections to Exploit Composition}
In the basic idea of the approach, linear neurons can be replaced by exploiting the compositional structure of the feed forward architecture. Equation \ref{composition_exploit} demonstrates how the weights can be distributed to the next layer, which now additionally needs the input of the skipped layer as input. The implementation uses shortcut connections to connect inputs to non-adjacent layers.
Shortcut connections carry the weights theoretically distributed into the next layer. During inference, a shortcut connection is evaluated as soon as the input is known, cached until the layer to skip is evaluated, and added onto before applying the layer activation. The input size of the skip connection is equal to the input size of the layer aimed to skip, and the output size is equal to the output size of the layer we are skipping into. 
An activation function is not used as the connection is only used as a tool to transfer the linear information.
The weights are calculated as described in Equation \ref{composition_exploit}. For multiple linear neurons in a layer, all weights for the same input are added together.

A special case that needs to be considered are shortcut connections in consecutive layers.  The second and each following consecutive layer must carry weights for both the original layer and the output of the shortcut connection. 
The weights are updated in the same way as before.
The effects of those rapidly growing consecutive shortcut connections and how to mitigate them will be resolved later.

With the introduction of shortcut connections, inference has to be adapted accordingly. For the details, we refer to Algorithm \ref{algo:forward} in the appendix. 
%
The main changes are two further checks, 1, does a skip connection end here in the form of a cached shortcut connection output and 2, do we skip from this layer further so that we also need this layer's input to evaluate the shortcut connections.

\section{Experiments}
The evaluation of the compression is a multi-step procedure. After introducing the datasets and models, the occurrence of provable linear neurons is tested. Afterwards the practical prototype is evaluated and adjusted, such that it can finally be compared and combined with a different compression method. All details regarding the experimental setup are available in the accompanying GitHub repository.

\subsection{Datasets}
To evaluate the compression performance, two datasets along with a benchmarking collection are utilized. The first dataset is the \textbf{Titanic} classification data set provided by the "kaggle" website.\citep{titanic} The second is the image classification task \textbf{Fashion-MNIST} by the zalando research group.\citep{fashionMNST} Lastly, the \textbf{OpenML-CC18} benchmark will be used.\citep{Bischl.2021} The collection of 72 classification tasks allows validating the findings from single datasets. Due to technical reasons only 61 of those 72 tasks where used. 
\subsection{Models}
For the Titanic dataset, we use a standard fully connected network with six hidden layers of dimensions (25, 50, 100, 100, 100, 100) with ReLU activations in the hidden layers and a softmax activation for the output. This results in overall 37,702 parameters.

The input images from the FashionMNIST dataset are flattened to enable compatibility with a standard fully connected neural network. This model has six hidden layers of dimensions (1024, 1024, 512, 512, 256, 256) with ReLU activations in the hidden layers and a softmax activation for the output. The parameters here add up to 2,840,586.

For all the models on the OpenML-CC18 benchmark the same architecture is used. This architecture has 5 hidden layers, of dimensions (64, 128, 128, 256, 256) with ReLU activations and a softmax activation in the output. The parameters for each model then are 123,648 plus all parameters, which are depend on the input or output size.

\subsection{Provable Linear Neurons}
To examine the occurrence of provable linear neurons in a practical setting, all models on the OpenML benchmark are checked for the two required conditions.

Out of the 61 models, only 5 contain neurons where all weights and the bias are trained to be positive. Across these 5 models, a total of 18 such neurons are found. However, all of them reside in the first hidden layer, which receives external input that may include negative values. As a result, none of the models ultimately contain a provable linear neuron.
\subsection{Practical Model Compression}
For the practical setting, the exact algorithm conducted from the theoretical idea of linearity-based compression is applied to compress the models trained on the Titanic and the FashionMNIST dataset.
We start with the compression of neurons with activation rates of at least $1.0$ and further compress the model with the same activation rates by iteratively reducing the threshold by $0.05$.
All compressed models of a given dataset are evaluated using the same test data to ensure consistency.

\begin{figure}[!t]
    \centering
     \begin{subfigure}{0.24\linewidth}
      \centering
       \includegraphics[width=1\linewidth]{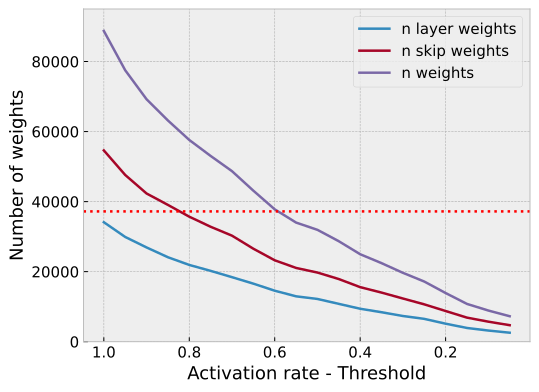}
        \caption{Titanic Weights}
       \label{fig:naiveWeights}
     \end{subfigure}
     \begin{subfigure}{0.24\linewidth}
      \centering
       \includegraphics[width=1\linewidth]{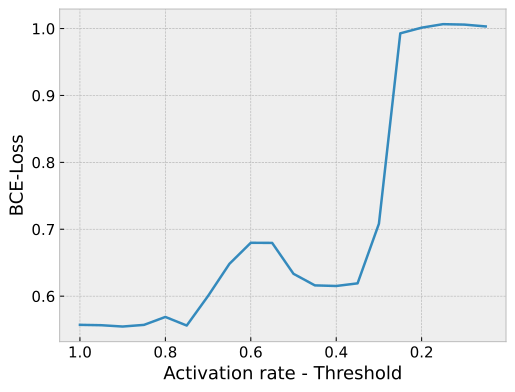}
        \caption{Titanic Loss}
       \label{fig:naiveLoss}
     \end{subfigure}
     \begin{subfigure}{0.24\linewidth}
         \centering
         \includegraphics[width=\linewidth]{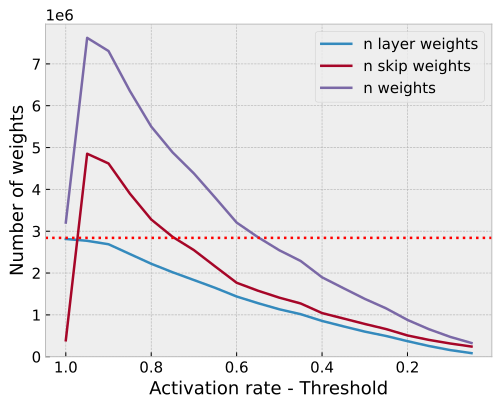}
         \caption{FashionMNIST Wghts.}
         \label{fig:naiveWeightsFashion}
     \end{subfigure}
     \begin{subfigure}{0.24\linewidth}
         \centering
         \includegraphics[width=\linewidth]{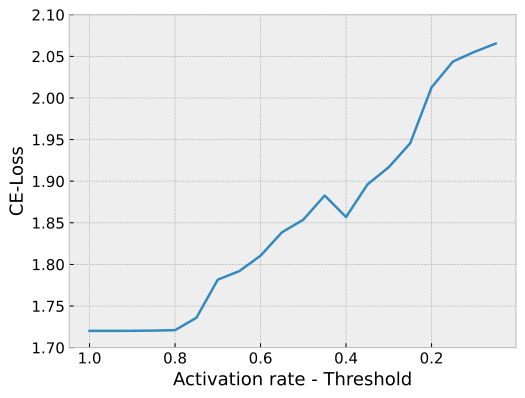}
         \caption{FashionMNIST Loss}
         \label{fig:naiveLossFashion}
     \end{subfigure}
  \caption{
  Compression results for FC networks on Titanic and FashionMNIST with 0.05 threshold steps and without layer threshold.
  Plots (a) and (c) show the number of weights (from bottom to top) in original layers, skip connections and in total. The dotted line marks the original model size. 
  Plots (b) and (d) show the corresponding loss for each activation rate threshold. Lowering the activation rate threshold leads to a decrease in the number of weights and an increase in the loss.
  }
  \label{fig:naiveComp}
\end{figure}

\begin{figure}[!t]
    \centering
     \begin{subfigure}{0.24\linewidth}
      \centering
       \includegraphics[width=\linewidth]{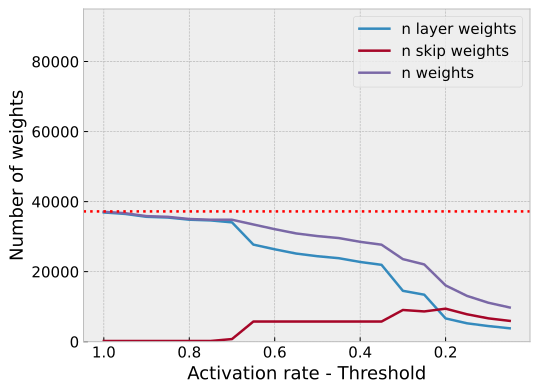}
       \caption{Titanic Weights}
       \label{fig:ImprovThresWeights}
     \end{subfigure}
     \begin{subfigure}{0.24\linewidth}
      \centering
       \includegraphics[width=\linewidth]{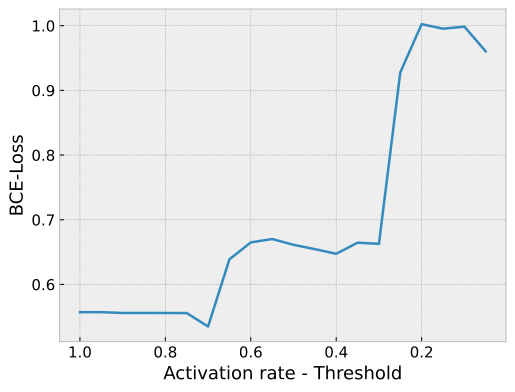}
       \caption{Titanic Loss}
       \label{fig:ImprovThresLoss}
     \end{subfigure}
     \medskip
     \begin{subfigure}{0.24\linewidth}
         \centering
         \includegraphics[width=\linewidth]{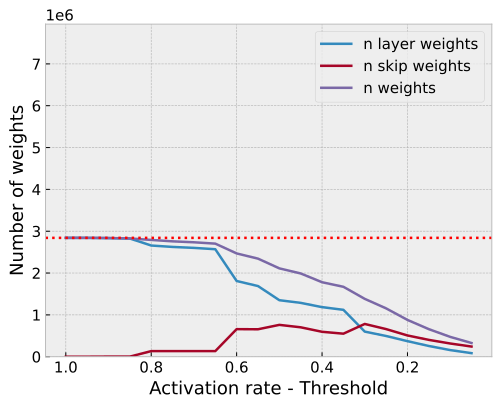}
         \caption{FashionMNIST Wghts.}
         \label{fig:ImprovThresWeightsFashion}
     \end{subfigure}
     \begin{subfigure}{0.24\linewidth}
         \centering
         \includegraphics[width=\linewidth]{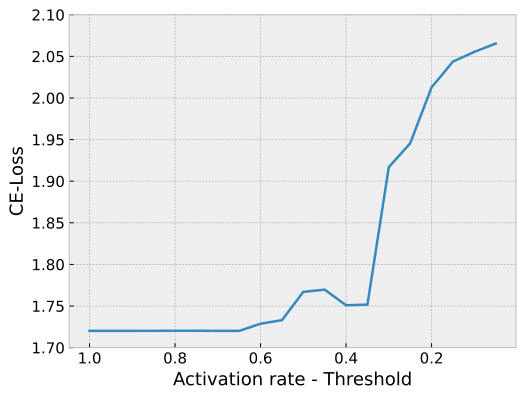}
         \caption{FashionMNIST Loss}
         \label{fig:ImprovThresLossFashion}
     \end{subfigure}
    \caption{
    Compression results for FC networks on Titanic and FashionMNIST with 0.05 threshold steps and optimal layer threshold.
    Plots (a) and (c) show the number of weights in original layers, skip connections and in total. The dotted line marks the original model size.
    Plots (b) and (d) show the corresponding loss for each activation rate threshold.
    Lowering the activation rate threshold leads to a strict decrease in number of weights in the model.
  }
    \label{fig:ImprovThresComp}
\end{figure}

Figure \ref{fig:naiveComp} shows how weights and loss change with different activation rate thresholds for both models. As expected, lower thresholds lead to higher loss, since more inputs incorrectly assume linearity. Due to many weights in the early skip connections, the original model size is only reached at an activation rate of 0.60 and 0.55, where the loss has already risen from 0.55 to 0.70 and from 1.72 to 1.85, respectively.
The two main reasons for the high number of initial weights are the number of linear neurons needed before the skip connection decreases the size and the occurrence of consecutive skip connections.
To address this, an additional threshold is introduced, requiring a minimum number of linear neurons before inserting a skip connection. This helps avoid small or consecutive skip connections.
A basic approach here would be an absolute value layer-threshold. As shown in Plot \ref{fig:absThresComp} in the appendix, this improves results but depends on a well-chosen threshold that suits all layer sizes.
The core aim of the layer-threshold is only to skip layers that reduce the overall size of the model. A more sophisticated approach can check exactly this. Using Inequation \ref{equ:compressionImprov} to analyze if a shortcut connection decreases or increases the number of weights, we can calculate the layer-threshold as,
\begin{align}
    \frac{wid_{i-1} wid_{i + 1}}{(wid_{i-1} + 1 + wid_{i+1})} \leq n_{linear} \label{equ:optLayerThresh} 
\end{align}
Note that existing shortcut connections are ignored, which could theoretically increase model size through consecutive shortcut connections.
For now, compression with this layer-threshold will achieve the desired properties. Figures \ref{fig:ImprovThresWeights} and \ref{fig:ImprovThresWeightsFashion} show that for all activation thresholds, the compressed model is smaller than both the original and any model compressed with a higher threshold. 
Loss remains flat above thresholds of 0.70 and 0.60, slightly increases down to 0.35, and rises sharply below that. The compression on both models have shown that using this compression, the model sizes can be reduced by ca. 10\% without change in performance and up to 1/3 with small increase in loss.   
The concrete choice of compression in practice depends on hardware settings, the subjective trade-off between size and performance, and the reasons why compression was desired in the first place. In general, this shows that linearity-based network compression can work and can compress networks to different sizes as a stand alone method.
\subsection{OpenML-CC18 Model Compression}
This subsection changes the perspective away from many compressions of a few models towards a few specified compressions for many models.
Each of the 61 OpenML-CC18 models are compressed to $75\%$, $50\%$ and $25\%$ of their original size. To do so, the original model is compressed with decreasing activation rate thresholds until the size first falls below the desired fraction of the original size. Using a test set neither used in training nor pruning the performance of the original and the three compressed models can be compared.
\begin{figure}[!t]
    \centering
    \begin{subfigure}{0.49\linewidth}
        \centering
        \includegraphics[width = 5 cm]{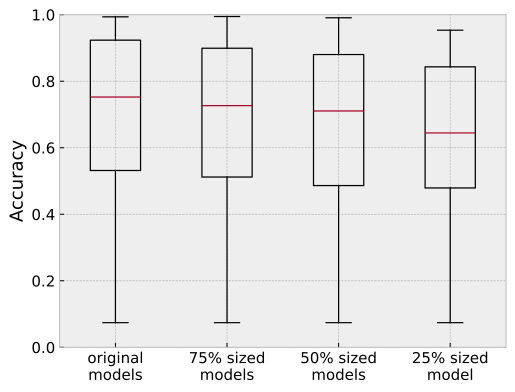}
        \caption{OpenML Boxplot}
    \end{subfigure}
    \begin{subfigure}{0.49\linewidth}
        \centering
        \includegraphics[width = 5 cm]{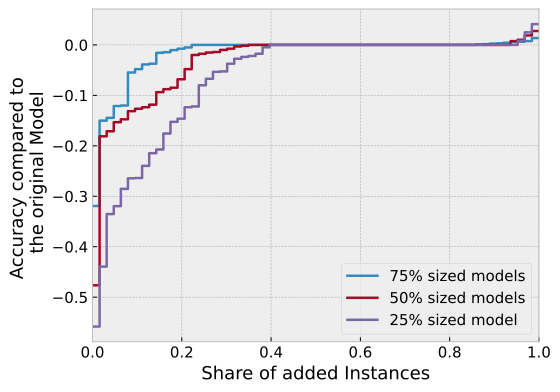}
        \caption{OpenML ECDF-Plot}
    \end{subfigure}
    \caption{Comparing the accuracy of compressing 61 classification tasks to 25\%, 50\% and 75\% of their original size. The majority of models is not effected by a compression to 1/4 of the original size.}
    \label{fig:OpenMLCompression}
\end{figure}
Figure \ref{fig:OpenMLCompression} depicts the slight decrease in accuracy in each compression.
While some individual models experience significant degradation, losing up to 0.5 in accuracy, show many compressed models more moderate drops: approximately 0.1 when compressed to 75\%, around 0.15 at 50\% compression, and up to 0.35 when reduced to 25\% of their original size.
However, the majority of models are uninfluenced by all three compressions. 
In summary, there are models that, on the one hand, can be heavily compressed without notable degradation, and on the other hand, also models that decline noteworthy already when compressed to 75\% of the original size. 
Those results are again validated across multiple architectures in the appendix in Plot \ref{fig:moreOpenML}. 
\subsection{Combination of Compression Techniques}
In addition to the proposed compression method, a simplified version of the activation-based algorithm from \citep{Ganguli.2024} is used for comparison and combination.
The fully connected model on the FashionMNIST dataset is compressed in two ways: first, using importance-based pruning alone, and second, by applying linearity-based compression to the result of the importance-based pruning, effectively combining both methods.
To apply the importance-based pruning, the model is further trained on the original training set while removing the 5\% lowest activated neurons after each epoch, provided the training accuracy is over a minimum of 0.70. The pruning process continues until the target compression level is achieved or a maximum of 15 epochs is reached.
The compression is evaluated using the three desired compression sizes 75\%, 60\%, and 50\% of the original model size.
Compression to 75\% of the original model size maintains accuracy at 0.741. At 60\% and 50\%, accuracy slightly drops to 0.736 and 0.720, respectively.
\begin{figure}[!t]
    \centering
     \begin{subfigure}{0.24\linewidth}
      \centering
       \includegraphics[width=1\linewidth]{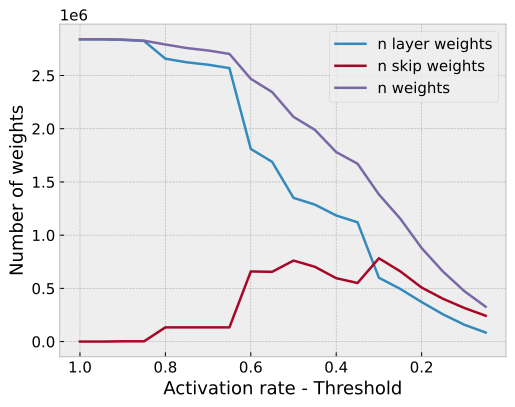}
       \caption{Unpru. Model Weights}
       \label{fig:unprunedFashionWeights}
     \end{subfigure}
     \begin{subfigure}{0.24\linewidth}
      \centering
       \includegraphics[width=1\linewidth]{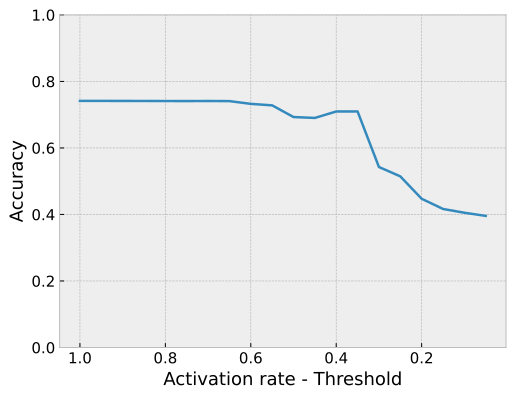}
       \caption{Unpru. Model Acc.}
       \label{fig:unprunedFashionLoss}
     \end{subfigure}
     \begin{subfigure}{0.24\linewidth}
      \centering
       \includegraphics[width=1\linewidth]{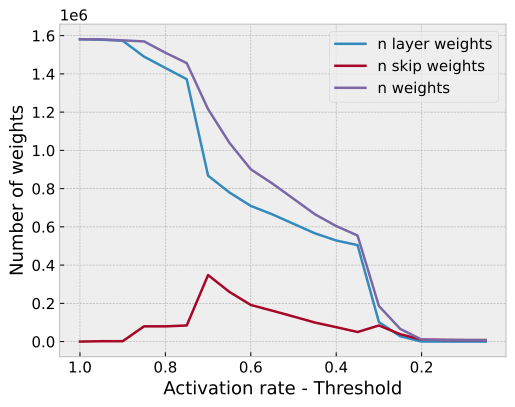}
       \caption{Prepru. Model Weights}
       \label{fig:preprunedFashionWeights}
     \end{subfigure}
     \begin{subfigure}{0.24\linewidth}
      \centering
       \includegraphics[width=1\linewidth]{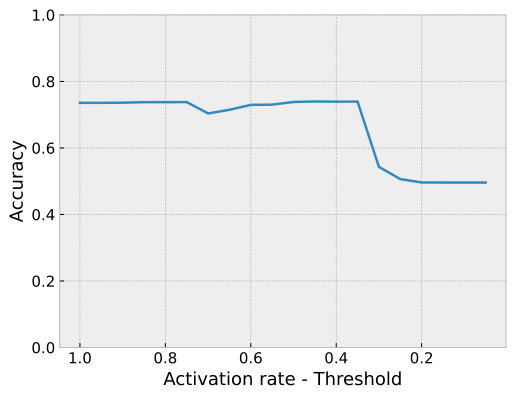}
       \caption{Prepru. Model Acc.}
       \label{fig:preprunedFashionLoss}
     \end{subfigure}
    \caption{
  Compression result for an unpruned and preprunded  FC network on the FashionMNIST dataset with a 0.05 threshold step size with optimal layer threshold.
  The plots (a) and (c) shows the number of weights in the original layers, the skip connections and the sum of them.
  The plots (b) and (d) shows the corresponding accuracy for each activation rate threshold. The loss runs nearly identical for both compressions.}
    \label{fig:unprunedFashion}
\end{figure}
Figure \ref{fig:unprunedFashion} shows the earlier linearity-based pruning results on the FashionMNIST model, using accuracy as the metric.
Linearity- and importance-based pruning achieve similar compression levels with comparable accuracy. While importance-based pruning performs slightly better, it benefits from refinement training, which could also enhance linearity-based compression.
Overall, the simple linearity-based method performs well against the established importance-based approach.

In the next step, the two approaches are combined to evaluate how much they might interfere with each other. In theory, interference should be minimal, as the two methods target different neurons for compression. This combination is achieved by applying the linearity-based compression to the model resulting from the importance-based compression to 60\% of the original model size. The 60\% compression level is selected as it retains nearly the original performance during pruning.
The result of this combined compression is plotted in Figure \ref{fig:unprunedFashion}. For activation rate thresholds above 0.75, the accuracy remains unchanged. These compressions reduce the model size from 1.60 million parameters to as low as 1.45 million parameters. For activation rate thresholds smaller than 0.75 but greater than 0.35, there is a slight decrease in accuracy. This minor performance loss allows the model size to be reduced to approximately 0.6 million parameters, effectively saving around one million parameters in the already pruned model.
The comparison with the linearity-based compression of the unpruned model shows that, in both cases, approximately one million parameters can be removed with little change in accuracy independent of the previously applied compression. 
This reinforces the theory that there is little interference between the two approaches of compressions. Consequently, combining linearity-based compression with other techniques enhances overall compression, as demonstrated in this experiment.

\section{Related Work}
The most common approach currently used is to estimate the importance of all neurons or components and remove the least important.
ConvNets\citep{ConvNets} present a basic importance-based approach, where filters in convolutional neural networks are pruned based on the sum of their weights. A similar, straightforward method is presented by Ganguli et al.\citep{Ganguli.2024}, which employs a pruning set to identify the least activated neurons during inference. It is the same analysis as in our work just focusing on least active neurons.
Other works use more sophisticated methods and greater computational resources to identify unimportance. So compare Molchanov et al.\citep{Molchanov.2019} the current model to alternatives missing single neurons, for which they want to compute the importance. "Neural network relief"\citep{Dekhovich.2024} prunes weights based on their average contribution to the neurons output.
It is also common to consider not only a component in isolation, but rather its role in the model. 
For instance, both ThiNet\citep{Luo.2017} and NISP\citep{Yu.2018} define a neuron's importance based on its contribution to reconstructing specific layers of the network. 
Most of the above-mentioned approaches include an optional refinement step, but some even more sophisticated approaches also use learning to identify unimportance. This way utilize Lin et al.\citep{Lin.2019} adversarial learning for pruning or Cheng et al.\citep{Cheng.2023} pruning together with the network architecture search task.

The second angle to approach pruning are the similarity-based methods , which exploit redundancy within the model to reduce its size. In these methods, identical or highly similar components are combined and represented by the same weights. Mukarrama et al. \citep{Mukarrama.2020} present a basic implementation of the general idea, using the cosine similarity between vectorized filters to measure redundancy. 
Again, also more advanced methods evolved to identify redundancy. The work of Li et al. \citep{Li.2019} for example uses the k-means algorithm to find clusters of filters, which are replaced by their centroids. Additionally, both Centripetal SGD\citep{Ding.2019} and NoiseOut\citep{noiseOut} regularize the training towards similarity of the components.

To achieve state-of-the-art compression results, several compression methods are combined.\citep{deep_compression}
This work introduces a new approach that complements existing methods by focusing on highly active components, typically regarded as highly important, and integrating them into other parts of the network.
The method presented in this work can be seen as a parallel to ConvNets and the work of Mukarrama, offering a practical and accessible implementation of the underlying concept. Best possible performance is not the highest priority, but rather to provide an understandable access point for further research.

\section{Discussion}
As a first result, we observe that across 61 models not a single provable linear neuron could be found. While full linearity is theoretically possible, and we could observe some cases in initial experiments, their occurrence appears to be too rare for meaningful exploitation to compress neural networks in practice.
Empirical linearity allowed the compression to be carried out and the size performance trade-off to be exploited.
Experiments on the Titanic and FashionMNIST models demonstrate that models can be compressed substantially while maintaining usable performance, with small reductions causing no noticeable loss. At the same time, small compressions can often be made without any noticeable change in loss. The OpenML compressions validate those findings and suggest that the compression success highly depends on the specific task. The majority of OpenML models could be compressed to less then 1/4 with no loss in performance.
From the optimal layer-threshold formula \ref{equ:optLayerThresh}, compression is most effective with varying layer sizes. When layer sizes are constant, at least half the neurons must be linear for meaningful compression. This is reflected in empirical results: deeper OpenML models with uniform layer sizes show weaker compression.
Compared to a basic existing method, the linearity-based approach performs competitively. Combining both techniques yields results supporting the theory that they interfere minimally.
Looking at it in the entirety, we can see mixed results for different circumstances. In some models, a majority of the weights can be compressed using linearity-based compression alone. In others, small compressions already decrease the performance to uselessness.
Nevertheless, can also be recorded that this simple implementation does already work and hence the linearity-based network compression has a justification for existence.

Yet, there only have been considered fully connected networks with ReLU activation functions in the hidden layers. In future works however, this concept can easily also be transferred to other architectures and activation functions.
Regardless of the specific model architecture, any component that includes fully connected layers can be compressed independently, as demonstrated here. Otherwise, linearity can be discovered and exploited in other architectures as well, potentially with the aid of further approximations. While transformers often utilize shallow fully connected blocks, where this approach is not promising, are there exceptions.  For instance, models like MobileBERT\citep{sun2020mobilebert} implement deeper fully connected blocks where linearity-based compression may offer meaningful benefits.
In terms of activation functions, the ReLU is a prominent and often used example but it is not the only option for our approach. The argument presented in this work extends to any piecewise-linear activation function. If a neuron consistently activates on the same linear segment, it would produce identical outputs using a linear activation function with the same slope and intercept as that segment. Non-piecewise-linear activation functions can also be approximated by a piecewise-linear activation function such that they can also be compressed.
This approximation would again introduce an error in prediction.
With these generalizations, the core algorithm presented here can be adapted and applied across a wide range of architectures and activation functions.
The method presented is just the most simple implementation of the theory. 
Future work should also enhance this compression through refinement training and better techniques for identifying linear neurons. Additionally, developing new compression strategies inspired by the core idea of removing linear neurons should be task of future work.

Finally, this work should be viewed in the context of its limitations and the broader research field. The focus has been solely on the reduction of the number of parameters, while another aspect of network compression, the reduction of inference time, has not be considered. Without closer examination, it is evident that the more complex inference function in this algorithm, does require more steps as well as more temporary memory.
This paper also omits all kinds of linearity-breaking components such as batch normalizations, which hinder the application of the linearity-based compression. The presence of such components would certainly impair the compression. 
A lot of other limitations and potential solution approaches also were just outlined in the future work.
The position of this work is not in a continuation of the existing line of research, but it splits of from an earlier, simpler state and points to a new direction. This new branch of research runs parallel to the established approaches. Its relevance to the current state-of-the-art lies in the current trend of combining multiple compression strategies, where this new branch allows for additional compression.
The method presented is not as trivially applicable as many existing techniques. Especially for more complex architectures the room of compression is strongly restricted. As a result, this approach cannot be used as generally as others and is aimed only at specialized applications, where peak compression should be achieved. 
\section{Conclusion}
Neural networks are compositions of weighted functions. This work contributes a theoretical understanding of how the linearity of such functions can be exploited to reduce the number of parameters, and under what conditions these functions exhibit linear behavior.
Building on this theory, a prototype compression algorithm was developed that identifies and replaces linear neurons in the network with skip connections. This method achieves compression results comparable to basic existing techniques. The combination with a different technique does not worsen the respective compressions noticeable.

\small{
\bibliography{fina_bibtex}

\begin{thebibliography}{20}
\providecommand{\natexlab}[1]{#1}
\providecommand{\url}[1]{\texttt{#1}}
\expandafter\ifx\csname urlstyle\endcsname\relax
  \providecommand{\doi}[1]{doi: #1}\else
  \providecommand{\doi}{doi: \begingroup \urlstyle{rm}\Url}\fi

\bibitem[{Andrew G. Howard} et~al.(2017){Andrew G. Howard}, {Menglong Zhu}, {Bo
  Chen}, {Dmitry Kalenichenko}, and {Hartwig Adam}]{MobileNets}
{Andrew G. Howard}, {Menglong Zhu}, {Bo Chen}, {Dmitry Kalenichenko}, and
  {Hartwig Adam}.
\newblock \emph{MobileNets: Efficient Convolutional Neural Networks for Mobile
  Vision Applications}.
\newblock 2017.

\bibitem[Babaeizadeh et~al.(2016)Babaeizadeh, Smaragdis, and
  Campbell]{noiseOut}
Mohammad Babaeizadeh, Paris Smaragdis, and Roy~H. Campbell.
\newblock \emph{NoiseOut: A Simple Way to Prune Neural Networks}.
\newblock 2016.

\bibitem[Bischl et~al.(2017)Bischl, Casalicchio, Feurer, Gijsbers, Hutter,
  Lang, Mantovani, {van Rijn}, and Vanschoren]{Bischl.2021}
Bernd Bischl, Giuseppe Casalicchio, Matthias Feurer, Pieter Gijsbers, Frank
  Hutter, Michel Lang, Rafael~G. Mantovani, Jan~N. {van Rijn}, and Joaquin
  Vanschoren.
\newblock Openml benchmarking suites.
\newblock \emph{ArXiv preprint}, 2017.

\bibitem[Brown et~al.(2020)Brown, Mann, Ryder, Subbiah, Kaplan, Dhariwal,
  Neelakantan, Shyam, Sastry, Askell, Agarwal, Herbert{-}Voss, Krueger,
  Henighan, Child, Ramesh, Ziegler, Wu, Winter, Hesse, Chen, Sigler, Litwin,
  Gray, Chess, Clark, Berner, McCandlish, Radford, Sutskever, and Amodei]{GPT3}
Tom~B. Brown, Benjamin Mann, Nick Ryder, Melanie Subbiah, Jared Kaplan,
  Prafulla Dhariwal, Arvind Neelakantan, Pranav Shyam, Girish Sastry, Amanda
  Askell, Sandhini Agarwal, Ariel Herbert{-}Voss, Gretchen Krueger, Tom
  Henighan, Rewon Child, Aditya Ramesh, Daniel~M. Ziegler, Jeffrey Wu, Clemens
  Winter, Christopher Hesse, Mark Chen, Eric Sigler, Mateusz Litwin, Scott
  Gray, Benjamin Chess, Jack Clark, Christopher Berner, Sam McCandlish, Alec
  Radford, Ilya Sutskever, and Dario Amodei.
\newblock Language models are few-shot learners.
\newblock 2020.

\bibitem[Cheng et~al.(2023)Cheng, Wang, Ma, Wei, Alsaadi, and Liu]{Cheng.2023}
Hanjing Cheng, Zidong Wang, Lifeng Ma, Zhihui Wei, Fawaz~E. Alsaadi, and
  Xiaohui Liu.
\newblock Differentiable channel pruning guided via attention mechanism: a
  novel neural network pruning approach.
\newblock \emph{Complex {\&} Intelligent Systems}, \penalty0 (5):\penalty0
  5611--5624, 2023.

\bibitem[Cukierski(2012)]{titanic}
Will Cukierski.
\newblock Titanic - machine learning from disaster, 2012.
\newblock URL \url{https://kaggle.com/competitions/titanic}.

\bibitem[Dekhovich et~al.(2024)Dekhovich, Tax, Sluiter, and
  Bessa]{Dekhovich.2024}
Aleksandr Dekhovich, David M.~J. Tax, Marcel H.~F. Sluiter, and Miguel~A.
  Bessa.
\newblock Neural network relief: a pruning algorithm based on neural activity.
\newblock \emph{Machine Learning}, \penalty0 (5):\penalty0 2597--2618, 2024.
\newblock ISSN 1573-0565.

\bibitem[Ding et~al.(2019)Ding, Ding, Guo, and Han]{Ding.2019}
Xiaohan Ding, Guiguang Ding, Yuchen Guo, and Jungong Han.
\newblock Centripetal {SGD} for pruning very deep convolutional networks with
  complicated structure.
\newblock pages 4943--4953. Computer Vision Foundation / {IEEE}, 2019.

\bibitem[Ganguli and Chong(2024)]{Ganguli.2024}
Tushar Ganguli and Edwin K.~P. Chong.
\newblock Activation-based pruning of neural networks.
\newblock \emph{Algorithms}, \penalty0 (1):\penalty0 48, 2024.

\bibitem[Han et~al.(2015)Han, Mao, and Dally]{deep_compression}
Song Han, Huizi Mao, and William~J. Dally.
\newblock Deep compression: Compressing deep neural networks with pruning,
  trained quantization and huffman coding, 2015.

\bibitem[Li et~al.(2017)Li, Kadav, Durdanovic, Samet, and Graf]{ConvNets}
Hao Li, Asim Kadav, Igor Durdanovic, Hanan Samet, and Hans~Peter Graf.
\newblock Pruning filters for efficient convnets.
\newblock OpenReview.net, 2017.

\bibitem[Li et~al.(2019)Li, Zhu, and Sun]{Li.2019}
Lianqiang Li, Jie Zhu, and Ming-Ting Sun.
\newblock Deep learning based method for pruning deep neural networks.
\newblock In \emph{2019 IEEE International Conference on Multimedia {\&} Expo
  Workshops (ICMEW)}. IEEE, 2019.

\bibitem[Lin et~al.(2019)Lin, Ji, Yan, Zhang, Cao, Ye, Huang, and
  Doermann]{Lin.2019}
Shaohui Lin, Rongrong Ji, Chenqian Yan, Baochang Zhang, Liujuan Cao, Qixiang
  Ye, Feiyue Huang, and David~S. Doermann.
\newblock Towards optimal structured {CNN} pruning via generative adversarial
  learning.
\newblock pages 2790--2799. Computer Vision Foundation / {IEEE}, 2019.

\bibitem[Luo et~al.(2017)Luo, Wu, and Lin]{Luo.2017}
Jian{-}Hao Luo, Jianxin Wu, and Weiyao Lin.
\newblock Thinet: {A} filter level pruning method for deep neural network
  compression.
\newblock pages 5068--5076. {IEEE} Computer Society, 2017.

\bibitem[Molchanov et~al.(2019)Molchanov, Mallya, Tyree, Frosio, and
  Kautz]{Molchanov.2019}
Pavlo Molchanov, Arun Mallya, Stephen Tyree, Iuri Frosio, and Jan Kautz.
\newblock Importance estimation for neural network pruning.
\newblock pages 11264--11272. Computer Vision Foundation / {IEEE}, 2019.

\bibitem[Mukarrama et~al.(2020)Mukarrama, {Al Azad}, and {Raqib
  Mahmud}]{Mukarrama.2020}
Mayesha Mukarrama, Abul~Kalam {Al Azad}, and Khan {Raqib Mahmud}.
\newblock Neural network compression by filter similarity detection and
  visualization.
\newblock IEEE, 2020.

\bibitem[Neill(2020)]{Overview_network_compression}
James~O' Neill.
\newblock An overview of neural network compression, 2020.

\bibitem[Sun et~al.(2020)Sun, Yu, Song, Liu, Yang, and Zhou]{sun2020mobilebert}
Zhiqing Sun, Hongkun Yu, Xiaodan Song, Renjie Liu, Yiming Yang, and Denny Zhou.
\newblock Mobilebert: a compact task-agnostic bert for resource-limited
  devices.
\newblock \emph{arXiv preprint arXiv:2004.02984}, 2020.

\bibitem[Xiao et~al.(2017)Xiao, Rasul, and Vollgraf]{fashionMNST}
Han Xiao, Kashif Rasul, and Roland Vollgraf.
\newblock Fashion-mnist: a novel image dataset for benchmarking machine
  learning algorithms, 2017.

\bibitem[Yu et~al.(2018)Yu, Li, Chen, Lai, Morariu, Han, Gao, Lin, and
  Davis]{Yu.2018}
Ruichi Yu, Ang Li, Chun{-}Fu Chen, Jui{-}Hsin Lai, Vlad~I. Morariu, Xintong
  Han, Mingfei Gao, Ching{-}Yung Lin, and Larry~S. Davis.
\newblock {NISP:} pruning networks using neuron importance score propagation.
\newblock In \emph{2018 {IEEE} Conference on Computer Vision and Pattern
  Recognition, {CVPR} 2018, Salt Lake City, UT, USA, June 18-22, 2018}, pages
  9194--9203. {IEEE} Computer Society, 2018.

\end{thebibliography}
}

\newpage
\appendix

\section{Appendix / supplemental material}

\RestyleAlgo{boxruled}
\begin{algorithm}[!h]
\SetAlgoLined
$current\_skip\_output \leftarrow \text{None}$\;
$cumulative\_inputs \leftarrow \text{None}$\;

\For{$layer$ in $model.layers$}{
    \If{$current\_skip\_output$}{
        $x\_tmp \leftarrow layer(x) + current\_skip\_output$\;
        \If{$layer.shortcut\_connection$}{
            $cumulative\_inputs \leftarrow concatenated(x, cumulative\_inputs)$\;
            $current\_skip\_output \leftarrow layer.shortcut\_connection(cumulative\_inputs)$\;
        } \Else {
            $current\_skip\_output \leftarrow \text{None}$\;
            $cumulative\_inputs \leftarrow \text{None}$\;
        }
        $x \leftarrow x\_tmp$
    } \Else {
        \If{$layer.shortcut\_connection$}{
            $cumulative\_inputs \leftarrow x$\;
            $current\_skip\_output \leftarrow layer.shortcut\_connection(x)$\;
        }
        $x \leftarrow layer(x)$
    }
}
\Return{$x$}\;
\caption{Algorithm for Model Inference}
\label{algo:forward}
\end{algorithm}

\begin{figure}[!h]
    \centering
    \begin{subfigure}{0.49\linewidth}
        \includegraphics[width = 0.8\linewidth]{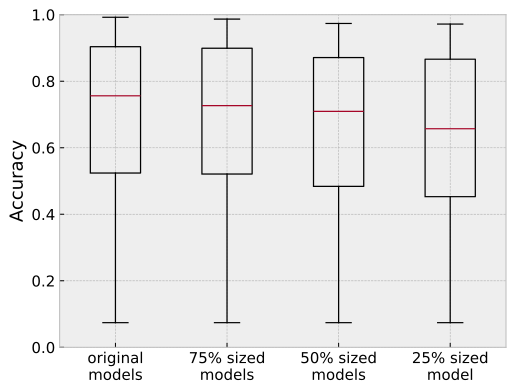}
        \caption{Half width}
        \label{fig:openMLhalfwidth}
    \end{subfigure}
    \begin{subfigure}{0.49\linewidth}
        \includegraphics[width = 0.8\linewidth]{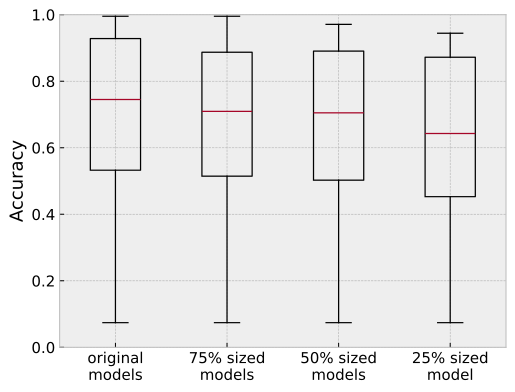}
        \caption{Double width}
        \label{fig:openMLdoublewidth}
    \end{subfigure}
    \begin{subfigure}{0.49\linewidth}
        \includegraphics[width = 0.8\linewidth]{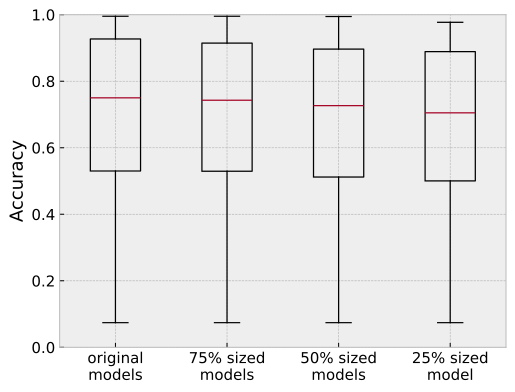}
        \caption{3 hidden layers}
        \label{fig:openMLsmalldepth}
    \end{subfigure}
    \begin{subfigure}{0.49\linewidth}
        \includegraphics[width = 0.8\linewidth]{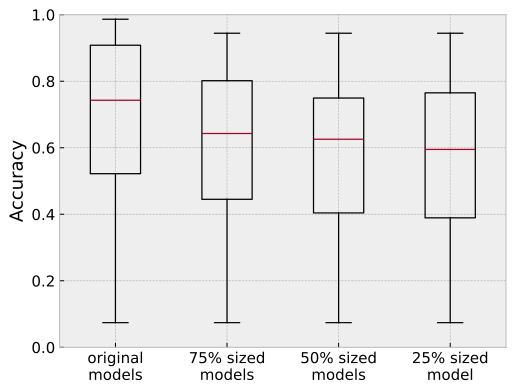}
        \caption{10 hidden layers}
        \label{fig:openMLbigdepth}
    \end{subfigure}
    \caption{Comparing the compression of 61 classification tasks to 25\%, 50\% and 75\% of their original size across different architectures. The compressions across all architectures yield similar results.}
    \label{fig:moreOpenML}
\end{figure}

\begin{figure}[!h]
    \centering
     \begin{subfigure}{0.475\linewidth}
      \centering
       \includegraphics[width=\linewidth]{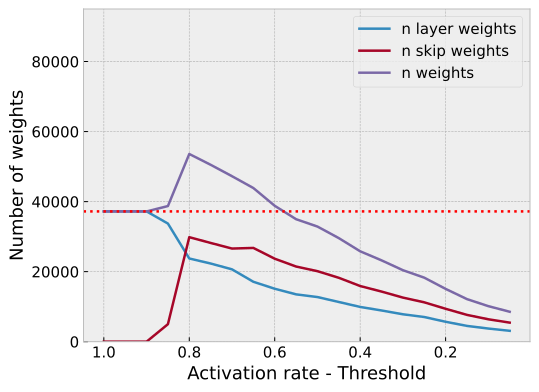}
       \caption{Titanic Weights}
       \label{fig:absThresWeights}
     \end{subfigure}
     \begin{subfigure}{0.475\linewidth}
      \centering
       \includegraphics[width=\linewidth]{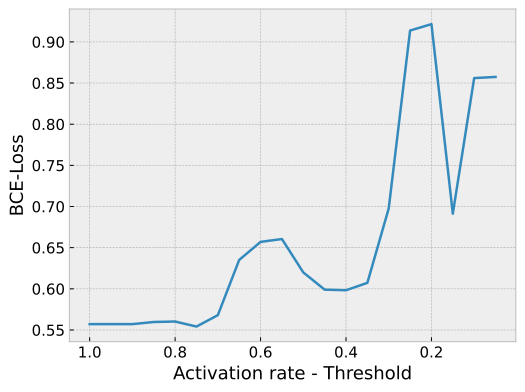}
       \caption{Titanic Loss}
       \label{fig:absThresLoss}
     \end{subfigure}
     \medskip
     \begin{subfigure}{0.475\linewidth}
         \centering
         \includegraphics[width=\linewidth]{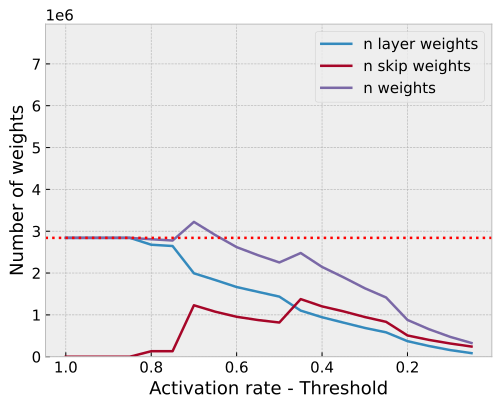}
         \caption{FashionMNIST Weights}
         \label{fig:absThresWeightsFashion}
     \end{subfigure}
     \begin{subfigure}{0.475\linewidth}
         \centering
         \includegraphics[width=\linewidth]{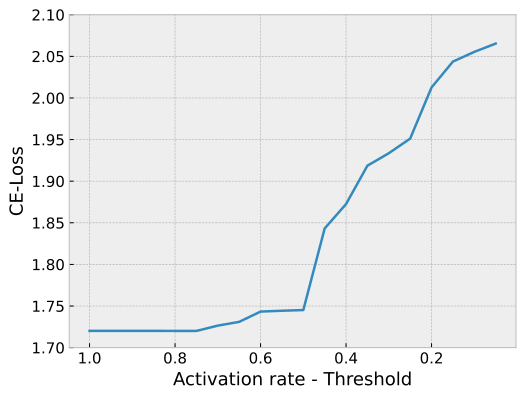}
         \caption{FashionMNIST Loss}
         \label{fig:absThresLossFasion}
     \end{subfigure}
    \caption{
    Compression results for FC networks on Titanic and FashionMNIST with 0.05 threshold steps and an absolute valued layer threshold.
    Plots (a) and (c) show the number of weights in original layers, skip connections and in total. The dotted line marks the original model size.
    Plots (b) and (d) show the corresponding loss for each activation rate threshold. The total number of weights decreases compared to the compression without layer threshold but still exceeds the original size of the models. 
  }
    \label{fig:absThresComp}
\end{figure}

\FloatBarrier

\end{document}